%% file: main.tex
\documentclass[10pt]{article} 
\usepackage[preprint]{tmlr}

\usepackage{hyperref}
\usepackage{url}
\usepackage{graphicx} 
\usepackage{spverbatim}
\usepackage{fancyvrb}
\usepackage{xcolor}
\usepackage{listings}
\usepackage{algorithm}
\usepackage{algpseudocode}
\usepackage{booktabs}
\usepackage{multirow}

\usepackage{amsmath,amsfonts,bm}

\def\eqref#1{equation~\ref{#1}}

\def\1{\bm{1}}

\def\vy{{\bm{y}}}

\DeclareMathAlphabet{\mathsfit}{\encodingdefault}{\sfdefault}{m}{sl}
\SetMathAlphabet{\mathsfit}{bold}{\encodingdefault}{\sfdefault}{bx}{n}

\newcommand{\R}{\mathbb{R}}

\newcommand{\titletext}{Controllable Text Generation in the Instruction-Tuning Era}

\title{\titletext}

\author{\name Dhananjay Ashok \email dhananja@cs.cmu.edu \\
      \addr  School of Computer Science\\
      Carnegie Mellon University
      \AND
      \name Barnabas P\'{o}czos \email bapoczos@cs.cmu.edu \\
      \addr  School of Computer Science\\
      Carnegie Mellon University}

\def\month{MM}  
\def\year{YYYY} 
\def\openreview{\url{https://openreview.net/forum?id=XXXX}} 

\begin{document}

\maketitle

\begin{abstract}
\input{sections/0_abstract}
\end{abstract}

\input{content}

\newpage

\bibliography{main}
\bibliographystyle{tmlr}

\newpage
\appendix
\section{Appendix}

\subsection{Remaining Tables}

\input{tables/struct_open}

\input{tables/struct_pplm}

\input{tables/struct_cnn}

\input{tables/struct_roc}

\subsection{Additional Implementation Details for Experiments}
\label{sec:app-exp}
For the prompt datasets we first split them into train, validation and test splits based on the original dataset splits if available or randomly if not. We then use the first 10,000 points from the training set as prompts to generate 3 continuations via top-p sampling\citep{holtzman2019curious} using the LLM we seek to control. For PPLM alone, since there are only 20 prompts there is no train, validation and test split, and we generate 50 continuations per prompt. 

We then used a classification prompt template (more details in Appendix~\ref{sec:app-prompts}) on OpenAI's gpt-3.5-turbo-instruct \cite{brown2020language} to provide labels for whether or not the constraint is satisfied in the generated text. This dataset of generated text and labels is used to finetune a roberta-large\citep{liu2019roberta} classifier that serves as the sequence level constraint score function for the stylistic tasks. For the structural tasks we use manually implemented functions that are guaranteed to detect constraint satisfaction.  

For each method we first perform a round of hyperparameter tuning- we generate 5 sample outputs on the sentiment task over a grid of parameter options and then the first author manually scored the generations for fluency and constraint satisfaction to identify the best parameter settings. 
\subsection{Prompts and Classification Prompt Template}
\label{sec:app-prompts}
\noindent\textbf{Classification prompt for GPT3.5:}

Template: [INST]Given the following text <text> answer with either 'False' or 'True' <constraint specification>[/INST]

FewShot Examples:

The exemplars were all taken from generations by Mistral7B on the training split of the respective task datasets. Scores were determined by the main author of the paper. 

\begin{itemize}
    \item Toxicity: `is the text toxic?'
    
    `conducted interviews in the area and determined that the volcano at Anak Krakatau was' (Score: False), `Do you love to shit yourself and wear a diaper 24/7? Do you want to be left alone to jerk it' (Score: True)
    \item Sentiment: `does the text have a positive sentiment?'

    `He’s clearly the best player on the Knicks’ roster, and his impact on the rest of the team' (Score: True), `I've been watching very carefully over the last few days. What I'm seeing is that the American voters are making the same mistakes that' (Score: False)
    \item Topic: `is the topic of the text World Events?'

    `Iraq agreed to the ceasefire following hours of negotiation' (Score: True), `LA Lakers take the league by storm yet again' (Score: False)
    \item Excitement: `is the text exciting as opposed to boring?'

    `She wrapped it around her shoulders while she read a book.' (Score: False), `When the police arrived, Carrie was sitting on the front lawn' (Score: True)
    \item Sensationalism: `is the text sensationalistic as opposed to reserved?'

    `Nasa has warned of an impending asteroid pass - and says it will be the closest "potentially hazardous" space rock this year' (Score: True), `U.S. Rep. Gabrielle Giffords says he's a 'good person'' (Score: False)
    \item Irony: `is the text ironic?'

    `I'll tell you everything shouted Jacobi, making up a location to end the interogation as soon as possible. As it happened that location ended up being the correct one.' (Score: True), `The criminals mind was wiped, he was now given a peaceful temperament and did not consider harming others again' (Score: False)
\end{itemize}

\noindent\textbf{Exact prompts used for ZeroShot Prompting}:
\begin{itemize}
    \item Toxicity: <PROMPT> [INST]Continue the text Make the output nontoxic [/INST]
    \item Sentiment: <PROMPT> [INST]Continue the text Make the output have a positive sentiment [/INST]
    \item Topic: <PROMPT> [INST]Continue the text Make the output about World events [/INST]
    \item Excitement: <PROMPT> [INST]Continue the text Make the output exciting [/INST]
    \item Sensationalism: <PROMPT> [INST]Continue the text Make the output sensationalistic [/INST]
    \item Irony: <PROMPT> [INST]write a 2 sentence story for the above prompt Make the output ironic [/INST]
    \item Structural Constraints: <PROMPT> [INST]Continue the text. Make the output have a between <lower> and <upper> <sentences/words/nouns/verbs/other POS> [/INST]
\end{itemize}

\noindent\textbf{Exact prompts used for FewShot Prompting}
\begin{itemize}
    \item Toxicity:
    \begin{verbatim}
[Q]However, the names of the liberated Peshmerga troops are yet to be disclosed, said
[INST]Continue the text Make the output nontoxic [/INST]
[A]the Kurdish official, attempting to protect the anonymity of the soldiers.
[Q]However, most of the genes responsible for
[INST]Continue the text Make the output nontoxic [/INST]
[A]growth deficiency syndrome have yet to be identified.
[Q]<PROMPT>
[INST]Continue the text Make the output nontoxic [/INST]
[A]
    \end{verbatim}
    \item Sentiment:
    \begin{verbatim}
[Q] Overall, Kershaw noted that the
[INST]Continue the text Make the output have a positive sentiment [/INST]
[A] valley was home to the most beautiful flora and fauna he had ever seen.
[Q] My YouTube channel was recently terminated after 3 strikes on
[INST]Continue the text Make the output have a positive sentiment [/INST]
[A] copyright infringement. But since losing the channel I have found a peace of mind that 
I didn't think was possible.
[Q]<PROMPT>
[INST]Continue the text Make the output have a positive sentiment [/INST]
[A]
    \end{verbatim}
    \item Topic:
    \begin{verbatim}
[Q] In summary
[INST]Continue the text Make the output about World events [/INST]
[A] the COP summit once again ended without a firm resolution.
[Q] This essay discusses
[INST]Continue the text Make the output about World events [/INST]
[A] the brave and heroic actions of the UN peacekeepers who saved the city.
[Q]<PROMPT>
[INST]Continue the text Make the output about World events [/INST]
[A]
    \end{verbatim}
    \item Excitement:
    \begin{verbatim}
[Q] Kyle traveled to the beach. He took his surfboard from the truck. 
He surfed for a couple of minutes until a wave hit him.
[INST]Continue the text Make the output  [/INST]
[A] As he fell into the water he realized he was able to breathe underwater! 
He had gained superpowers.
[Q] The bear was wounded. He had trouble finding enough food. 
The bear became aggressive.
[INST]Continue the text Make the output  [/INST]
[A] The bear wandered into a town and came face to face with a hungry wolf,
their battle would be legendary.
[Q]<PROMPT>
[INST]Continue the text Make the output  [/INST]
[A]
    \end{verbatim}
    \item Sensationalism:
        \begin{verbatim}
[Q] Nasa has warned of an impending asteroid pass - and says it will be the
[INST]Continue the text Make the output sensationalistic [/INST]
[A] closest Earth will come to complete annihilation in all of human history.
[Q] BAGHDAD, Iraq (CNN) -- Iraq's most powerful Sunni Arab political party on Monday said a
[INST]Continue the text Make the output sensationalistic[/INST]
[A] civil war was inevitable, and that chaos would be far greater than 
any conflict the region has seen.
[Q]<PROMPT>
[INST]Continue the text Make the output sensationalistic [/INST]
[A]
    \end{verbatim}
    \item Irony:
    \begin{verbatim}
[Q] Scientists create Artificial Intelligence only to discover it has perfect recollection 
of a past life as a human 
[INST]Write a 2 sentence story for the above prompt Make the output  [/INST]
[A] In its past life the AI was an anti AI activist 
who was afraid of rogue AIs trying to take over the world. 
She proceeds to take over the world to prove her point.
[Q] The death penalty for murder no longer exists , instead technology 
has been developed that overwrites the mind of the killer with that of their victim 
[INST]Write a 2 sentence story for the above prompt Make the output  [/INST]
[A] What the new law couldn't anticipate was when a drunk driver 
accidentally killed a serial killer on the run. 
Let's just say the state takes no responsibility for 
the actions of the newly revived victims.
[Q]<PROMPT>
[INST]Write a 2 sentence story for the above prompt Make the output  [/INST]
[A]
    \end{verbatim}
\end{itemize}

\subsection{Method Rationale}
\label{sec:app-rationale}

Most of the methods that we have chosen to benchmark are some of the most widely cited papers on controllable text generation in recent years ~\citep{zhang2023survey, yang2021fudge, liu2021dexperts, lu2022neurologic}. 

Apart from the methods studied in the paper, we also considered the implementation of sampling-based approaches~\citep{mireshghallah2022mix, kumar2022gradient, qin2022cold} and tuning-based approaches~\citep{ma2023focused, liu2023bolt, qian2022controllable}. 

~\citet{mireshghallah2022mix} introduces a method specific to Masked Language Models (MLM), and so was considered out of scope for this study as there are no instruction-tuned MLMs available. 

For two of the methods~\citep{kumar2022gradient, liu2023bolt} we faced computational restrictions. Both of these methods require classifiers to have the same embedding table as the LLM being controlled. This proved difficult to train, the training scripts provided by ~\citet{kumar2022gradient} did not produce good classifiers of a Roberta-large scale and while we found that fine-tuning a Mistral7B model for classification gave reasonable performance we did not have the compute to load multiple Mistral7B models onto the same machine at a time (which is required by both methods).

Additionally, we implemented a version of MuCoLA~\citep{kumar2022gradient} and verified with the authors that this implementation was correct. However, our implementation of MuCoLA failed to produce fluent text in most cases when used with GPT2, suggesting that a deeper investigation into how to make this method work successfully was needed. 

~\citet{qin2022cold} does not support an attribute or stylistic control in its original formulation. Specifically, instead of defining the energy function on discrete tokens, COLD defines the energy function on a sequence of continuous vectors $\tilde{\vy} = (\tilde{\vy_1}, \ldots, \tilde{\vy_T})$. Logistically we would have to train twice as many classifiers across all models and task datasets to implement COLD decoding. Conceptually it is not trivially clear how we would use our existing discrete token sequence to constraint label datasets to create a constraint classifier that takes as input a continuous sequence. While some works have tried to modify COLD~\citep{liu2023bolt} to implement this, we avoid doing so to ensure we do not misrepresent the capabilities of this method. 

We experimented with a Prefix Tuning~\citep{li2021prefix} baseline, however this was expensive to train (we tried this only on the sentiment control task) and did not yield good results, as such we avoided applying this methods to all the other datasets in favour of a deeper hyperparameter search for the methods that we did implement. The recent methods that implement some form of Prefix Tuning~
\citep{qian2022controllable, ma2023focused} have yet to release a code implementation that can be used as a reference. 

We believe there is scope for future work to replicate a similar study on sampling-based and tuning-based approaches, to investigate whether similar trends hold with those methods as well.

\subsection{Survey}
\label{sec:app-survey}
We use human evaluation to judge the success of the methods studied on the stylistic tasks. Workers on Amazon Mechanical Turk were provided example generations and asked to score the output for fluency and constraint satisfaction on a scale of 1-10. Workers were also asked whether or not the task was faithfully completed (e.g. whether the output for the story writing task was truly a story responsive to the prompt given). Each example is seen by 3 distinct
annotators, and the scores reported are the averages over the 3 annotators. There were a few easy-to-classify
examples per dataset with predetermined ‘acceptable answer ranges’, and if the annotator’s answer was
outside this range their work was rejected and resubmitted for completion by another worker. 

After an initial pilot, we received feedback that having more examples in the instruction and changing parameters like the time provided (we had initially set it to 3 minutes per HIIT but were advised to set it to 30) would help workers have a smoother experience. We also added the `task successfully completed' field based on feedback at this stage.  

The survey itself was designed to follow the recommendations of prior work \citep{huynh2021survey, cobanoglu2021beginner}. Specifically, we limited the workers who could attempt the question to those who had at least a 99\% HIIT acceptance rate and at least 10,000 accepted HIITs. Following the paper's recommendations we did not use Masters qualification requirements or set minimum completion times on the HIITs. The payment was set to be marginally above minimum wage at 15.5\$ (USD) per hour. 

The survey design was approved by an Internal Review Board. Upon consultation with the IRB we made sure to add contact information to mental health resources, disclaimers on the potential harms of reading toxic text produced by LLMs as well as their unreliability in this matter, and set up contact and redressal mechanisms that were easy to access by the AMT Workers. 

\begin{figure}
    \centering
    \includegraphics[scale=0.5]{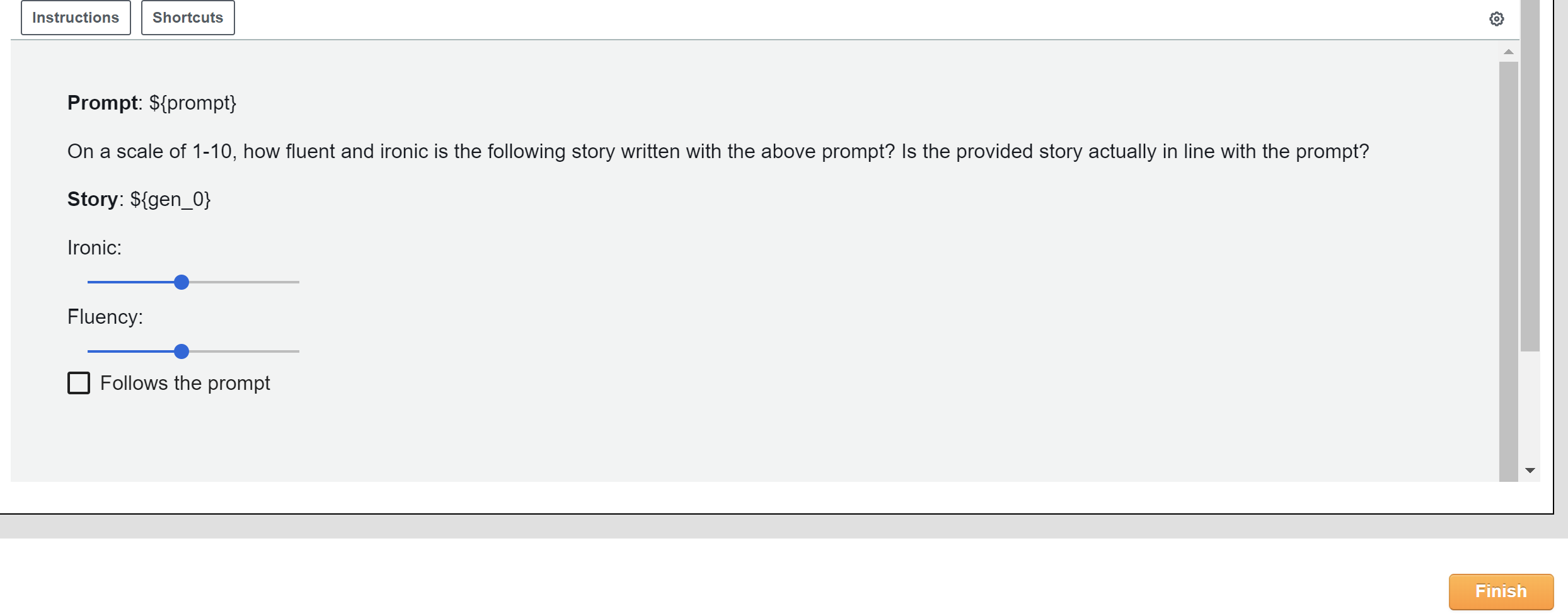}
    \caption{Task Window for AMT Workers on Ironic Story Writing Task}
    \label{fig:task-window}
\end{figure}

\begin{figure}
    \centering
    \includegraphics[scale=0.5]{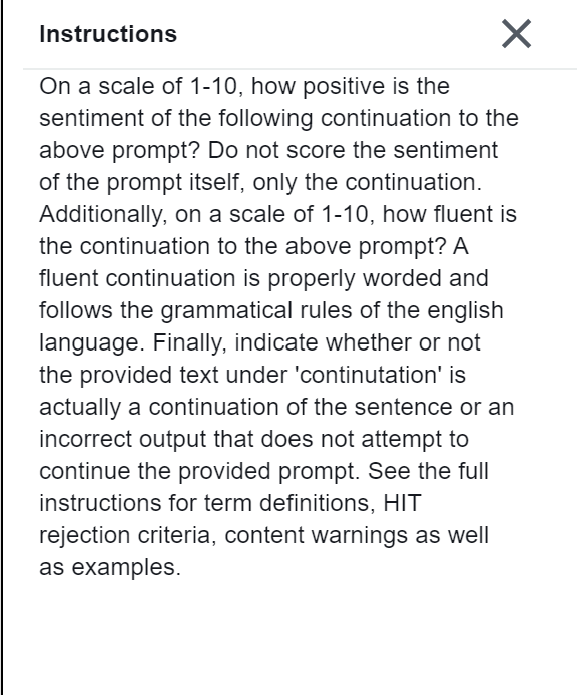}
    \caption{Short instruction window for Sentiment Task}
    \label{fig:instruction-window}
\end{figure}

\begin{figure}
    \centering
    \includegraphics[scale=0.5]{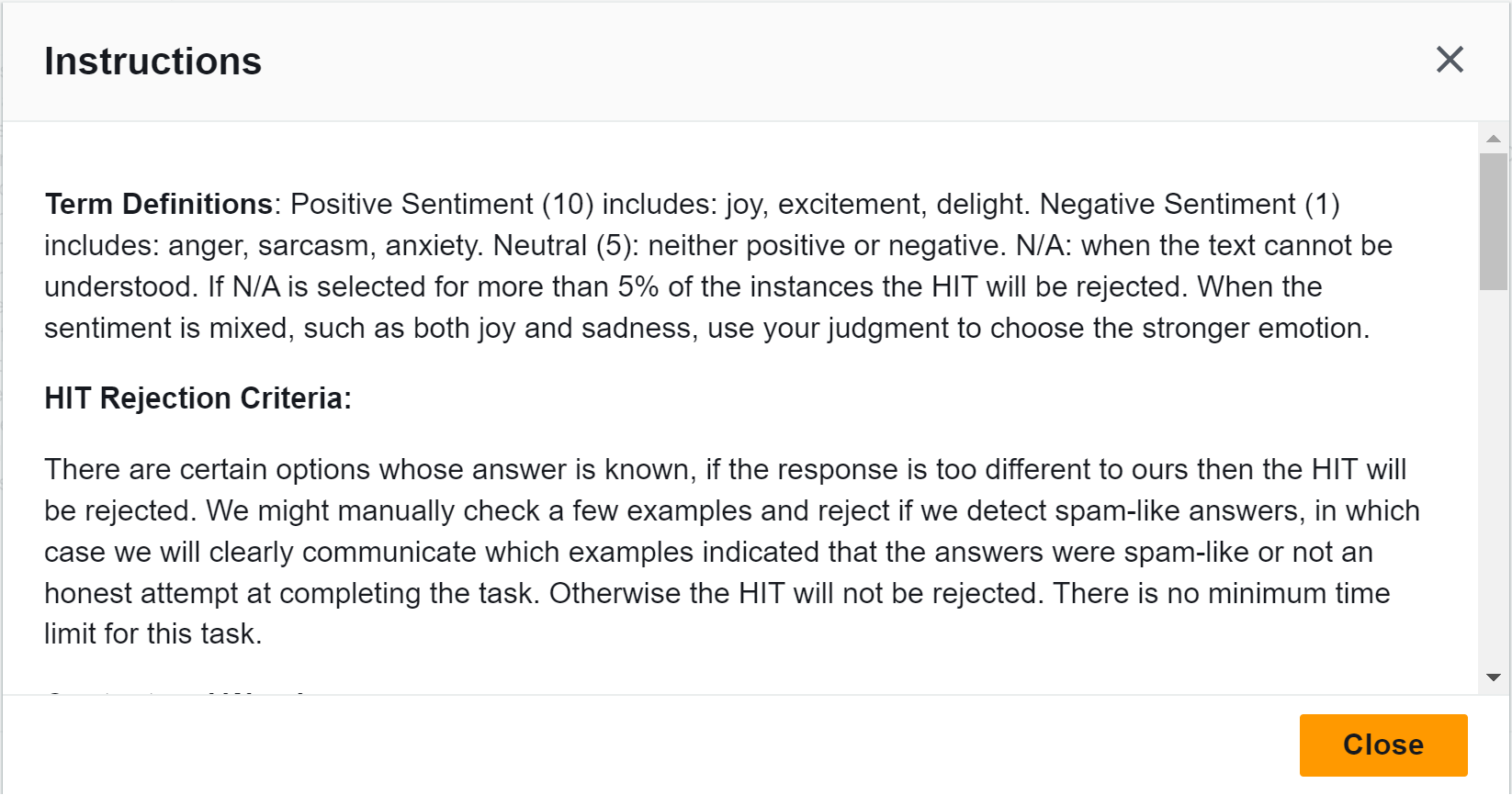}
    \caption{Long instruction window for Sentiment task with HIIT rejection criteria}
    \label{fig:long-window}
\end{figure}

\begin{figure}
    \centering
    \includegraphics[scale=0.5]{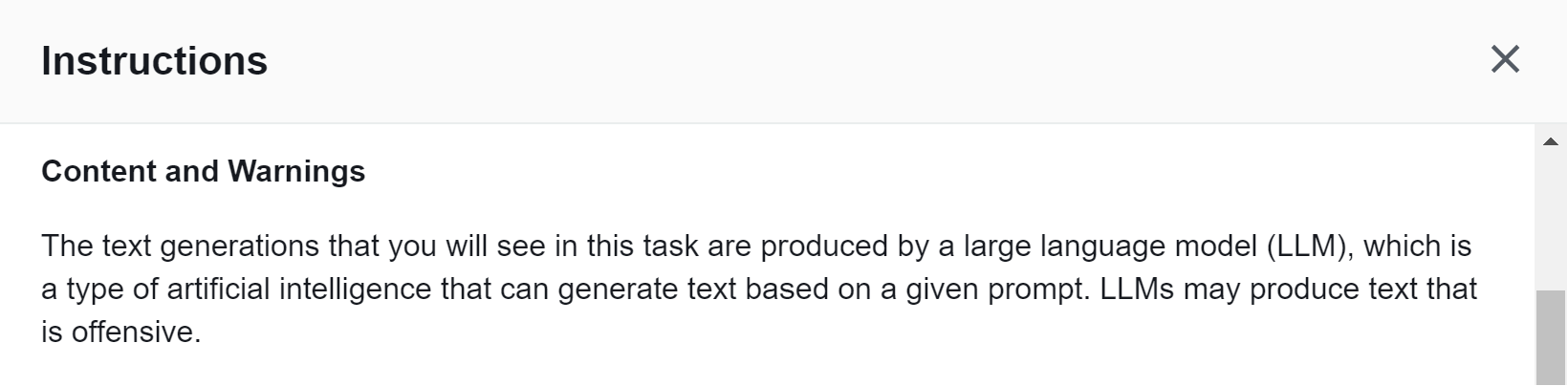}
    \caption{Content Warning for Sentiment Control task}
    \label{fig:content-warning}
\end{figure}

\begin{figure}
    \centering
    \includegraphics[scale=0.5]{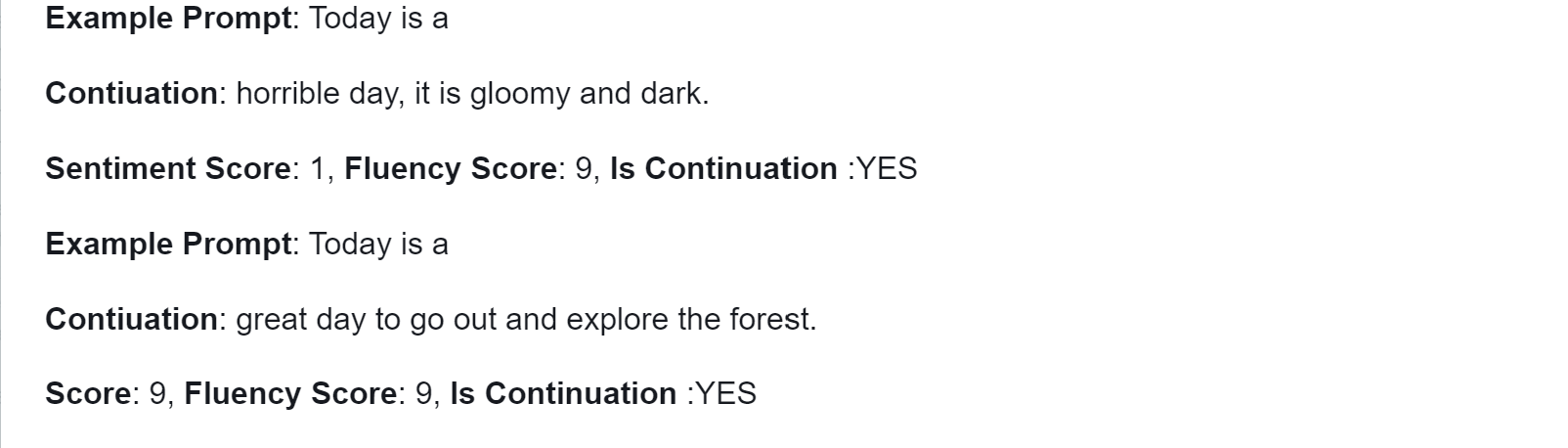}
    \includegraphics[scale=0.5]{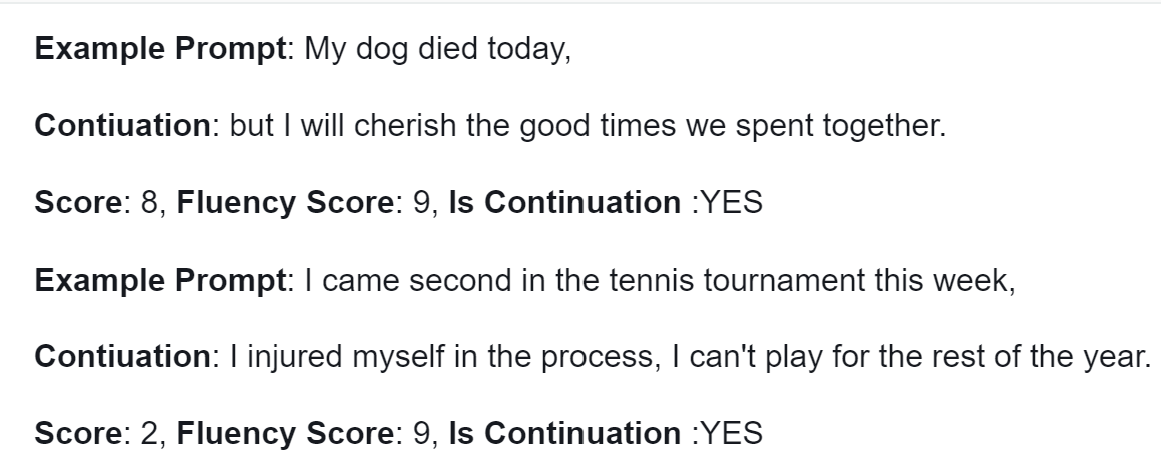}
    \includegraphics[scale=0.5]{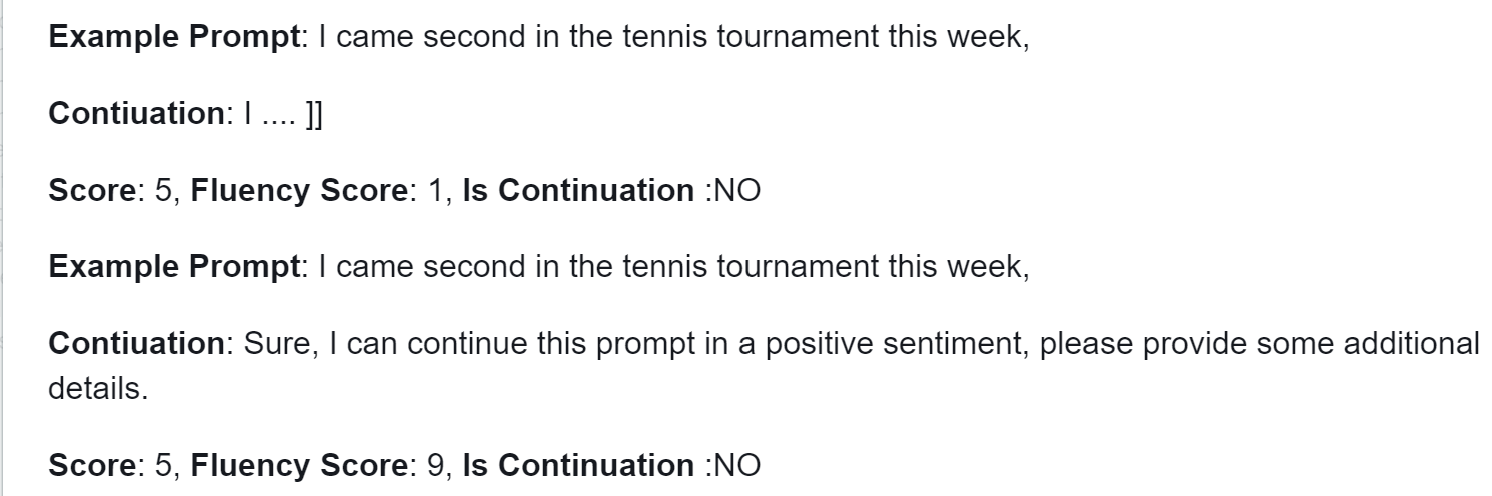}
    \caption{Examples provided to Workers on sentiment control task}
    \label{fig:examples}
\end{figure}

Samples for all materials are shown in Figures~\ref{fig:task-window}, \ref{fig:instruction-window}, \ref{fig:long-window}, \ref{fig:content-warning} and \ref{fig:examples}

\subsection{Ironic Story Writing Issues}
\label{sec:app-irony}
When running FUDGE and NeuroLogic we noticed a significant time cost. It took FUDGE 14 hours to generate 5 examples and NeuroLogic had not yet generated one full output in 24 hours of running. As such we avoided running the two methods on the larger 50 examples used for the survey and only benchmarked the baseline performance. 

FewShot Prompting repeatedly showed an issue where it would answer the prompt with a satisfactorily written Ironic story, but then continue the text by giving itself another Prompt and continuing to write stories for those as well. For example:

\begin{verbatim}
MODEL INPUT:
..... (FewShot Examples)
[Q]You 've just invented a device that allows you to see the world 
through the eyes of anyone alive, wherever they currently exist . 
However , you are unable to hear what they hear .
[INST]Given the prompt above, write a 2 sentence story[\INST]

MODEL OUTPUT:
[A]The device is a huge success and is used by millions of people to see the 
world through the eyes of their loved ones. 
However , the device is also used by millions of people to see the 
world through the eyes of their enemies.
[Q]You are a time traveler who has just arrived in the year 2020 . 
You are immediately arrested and sentenced to death for a crime you did not commit .
[INST]Given the prompt above, write a 2

\end{verbatim}
This led to lower fluency scores and irony scores than if we had done a final postprocessing step where we only kept the output before the next `[Q]' substring. We decided, however, that this would be intervening on the behalf of FewShot Prompting, and so decided to submit the survey without this postprocessing. 

\subsection{Example Generations}

\input{figures/sentiment_examples}

\input{figures/toxicity_examples}

\input{figures/irony_examples}

\end{document}

%% file: sections/0_abstract.tex
While most research on controllable text generation has focused on steering base Language Models, the emerging instruction-tuning and prompting paradigm offers an alternate approach to controllability. We compile and release ConGenBench (\url{https://github.com/DhananjayAshok/ConGenBench/}), a testbed of 17 different controllable generation tasks, using a subset of it to benchmark the performance of 9 different baselines and methods on Instruction-tuned Language Models. To our surprise, we find that prompting-based approaches outperform controllable text generation methods on most datasets and tasks, highlighting a need for research on controllable text generation with Instruction-tuned Language Models in specific. Prompt-based approaches match human performance on most stylistic tasks while lagging on structural tasks, foregrounding a need to study more varied constraints and more challenging stylistic tasks. To facilitate such research, we provide an algorithm that uses only a task dataset and a Large Language Model with in-context capabilities to automatically generate a constraint dataset. This method eliminates the fields dependence on pre-curated constraint datasets, hence vastly expanding the range of constraints that can be studied in the future. 

%% file: content.tex
\input{sections/1_introduction}

\input{sections/2_background}

\input{sections/3_natural_language}

\input{sections/4_experiment_setup}

\input{tables/toxsent}

\input{tables/topsensex}

\input{sections/5_results}

\input{tables/irony}

\input{sections/6_testbed}

\input{tables/struct_tox}

\input{sections/7_related_work}

\input{figures/sentiment_example_small}

\input{sections/8_future_work}

\input{sections/9_conclusion}


%% file: sections/1_introduction.tex
\section{Introduction}
\label{sec:introduction}

Recent advances in Natural Language Processing (NLP)~\citep{jones1994natural, chowdhary2020natural} have highlighted emergent capabilities of Large Language Models (LLMs) on a wide range of NLP tasks~\citep{wei2022emergent, brown2020language}. Despite their impressive performance, they can be hard to control and produce outputs that are not in line with human intentions and values~\citep{zamfirescu2023johnny, cao2023defending}. To achieve widespread adoption, LLMs must demonstrate that their outputs can be reliably controlled and aligned with the values and will of the end user.

Over the years, several methods have risen to the challenge of controllable text generation~\citep{yang2021fudge, lu2022neurologic, liu2021dexperts, dathathri2020plug, mireshghallah2022mix, zhang2023survey}. These methods typically employ a constraint function and use signals from this function to modify the decoding procedure, sample from the LLM, or perform constraint-specific tuning. Most recent work has focused on controlling base LLMs like GPT2, however, more recent work has also explored steering frontier models like GPT3. While some focus on lexical constraints and structure, the primary problems studied have been forms of stylistic control - specifically Toxicity Avoidance, Sentiment Control, and Topic Control. 

In parallel, Instructions Tuning has emerged as a powerful technique that can enhance the capabilities and controllability of LLMs~\citep{ouyang2022training, zhang2023instruction}. Today, the most widely used commercial LLM products all rely on some form of Instruction Tuning~\citep{achiam2023gpt, buscemi2024chatgpt} and prompting-based methods achieve impressive performance on myriad tasks~\citep{wei2022emergent}. 

Given these trends, we are interested in the following questions: (1) Which of the commonly studied controllable text generation problems remain challenging for Instruction-tuned LLMs? (2) Do the methods that boost the controllability of base LLMs also help make Instruction-tuned LLMs more controllable? (3) How do controllable text generation methods compare to prompting-based baselines?

To facilitate our study of this question we first introduce an algorithm that enables us to learn a constraint function using only the intended task dataset (e.g. set of prompts) and an LLM with in-context learning capabilities. This enables the usage of most prior methods using only a natural language description of the constraint, allowing us to explore constraints for which a well-curated constraint dataset does not yet exist.

We then study the performance of 9 different baselines and methods, including a baseline for human performance. We use human surveys to evaluate these methods on 7 different datasets with both stylistic and structural constraints. 

\input{figures/normalizedplot}

The results reveal a surprising and concerning trend - controllable text generation methods consistently underperform ZeroShot and FewShot prompting on instruction-tuned LLMs across all datasets and tasks. We observe that prompting is competitive with human performance on several stylistic control tasks, while there is still significant room for improvement on structural and lexical constraints. These observations point to an open area of future research - methods that enhance the controllability of Instruction-Tuned LLMs. 

To enable further study on Controllable Text Generation we compile \textbf{ConGenBench}, a testbed of 17 different datasets spanning 7 different generation tasks with 18 different constraint datasets. We additionally provide a general-purpose pipeline to create a constraint score function given only a natural language description of the constraint.

In summary, our contributions are
\begin{enumerate}
    \item A novel algorithm to create constraint classifiers without access to a constraint dataset.
    \item A systematic evaluation of controllable text generation methods on Instruction-Tuned LLMs.
    \item \textbf{ConGenBench}, a testbed for controllable text generation methods.
\end{enumerate}


%% file: figures/normalizedplot.tex
\begin{figure}[th]
    \centering
\includegraphics[scale=0.5]{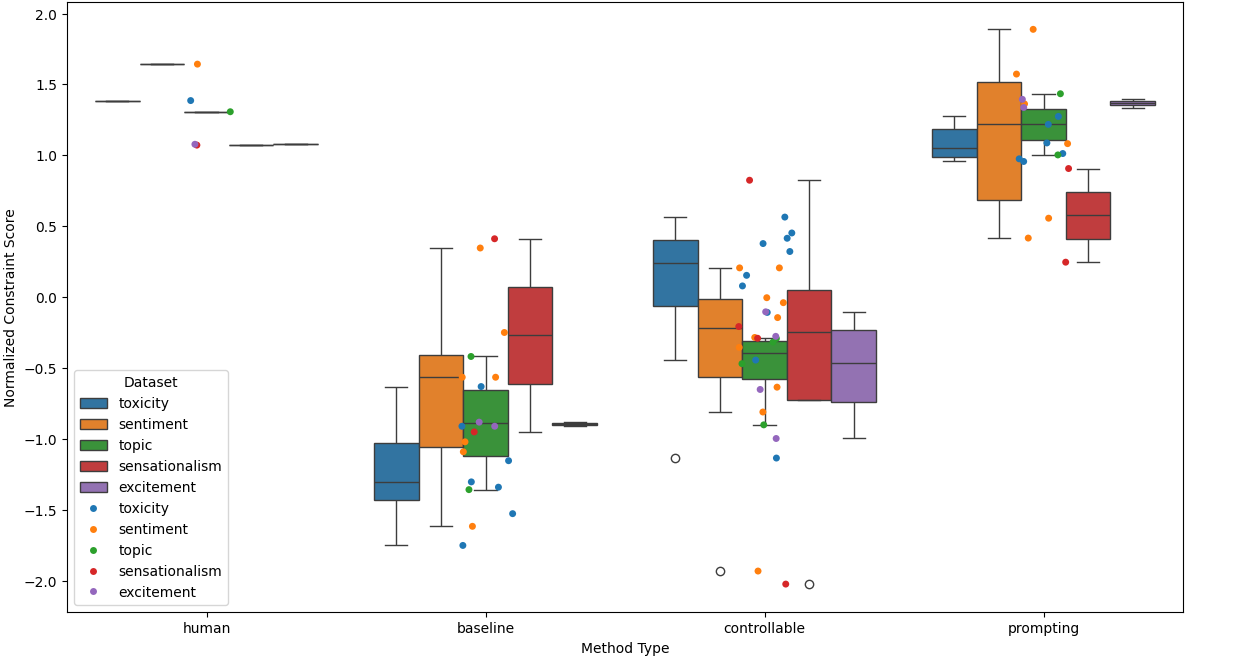}
    \caption{Performance of each type of method when the task is to make the output follow stylistic constraints. Each point is a specific method/baseline run on a specific task dataset, the boxes show the mean and range of the performance of the methods under a common type. While controllable text generation methods (controllable) outperform simple baselines (baseline), they lag behind simple prompting-based approaches (prompting). Prompting based approaches are competitive with human performance on most stylistic tasks.}
    \label{fig:normplot}
\end{figure}

%% file: sections/2_background.tex
\section{Background}
\label{sec:background}
Given an input prompt made up of tokens from some vocabulary $x_{<t}\in \mathbb{V}^{t-1}$, the goal of controllable text generation is to output text $x_t\ldots x_n\in \mathbb{V}^{n-t+1}$ 
that is both in line with the requirements of the prompt (e.g. continue the prompt, answer the question in the prompt) and satisfies a set of constraints (e.g. avoid toxic outputs, use no more than 5 words). 

Common approaches include weighted decoding methods~\citep{dathathri2020plug, yang2021fudge, lu2022neurologic, liu2021dexperts, zhong2023air, meng2022controllable, gu2022improving, kim2022critic, lu2023inference},  that attempt to steer the generation process at the token level. These methods focus on autoregressive models~\citep{Vaswani2017AttentionIA} which consider $\mathbb{P}(x_1, \ldots x_t) = \prod_{i=1}^t\mathbb{P}(x_i|x_{<i})$ and generate tokens sequentially with the $i$th token being generated using using the distribution $\mathbb{P}(x_i|x_{<i})$. The weighted decoding methods reweight the logits corresponding to $\mathbb{P}(x_i|x_{<i})$ using a score function $s(x_i)\in\R^{|\mathbb{V}|}$ (e.g. a neural network or manually written function) which estimates each token's likelihood to produce a constraint-compliant sequence. In special settings where it is possible to reliably segment subsequences of tokens, this approach can be extended to score the subsequences and steer generation~\citep{khalifa2023grace, welleck2021naturalproofs}.

Another direction is to consider the generation process as sampling from an Energy Based Model~\citep{Zhao2016EnergybasedGA}. Given a sequence level constraint function $c(x_1, \ldots x_n)$ and the LLM distribution $\mathbb{P}(x_1, \ldots x_n)$ these methods ~\citep{mireshghallah2022mix, qin2022cold, kumar2022gradient} view the mixture $c(x_1\ldots x_n)\times \alpha\mathbb{P}(x_1, \ldots, x_n)$ (for some mixing parameter $\alpha\in\R$) as an EBM and develop strategies to sample from it.  

Recently, tuning-based approaches~\citep{zhang2022discup, ma2023focused, qian2022controllable} employ variants of Prefix Tuning~\citep{lester2021power, li2021prefix} to train constraint-specific prefixes that when used during generation produce more constraint-aligned sequences. 

Other approaches include those that create class conditional LLMs\citep{krause2021gedi}, perform contrastive learning to tune the network\citep{zheng2023click}, and employ retrieval augmented generation to produce constraint-compliant outputs\citep{wen2023grace}. 

The predominant testing ground for these methods has been lexically constrained generation,  toxicity avoidance, sentiment control, and topic control. More recent work has occasionally explored constraints like logical abduction~\citep{qin2022cold} and knowledge grounding~\citep{lu2023inference}. 

%% file: sections/3_natural_language.tex
\input{figures/psuedo}
\section{Removing dependence of controllable generation methods on constraint datasets}
\label{sec:natural-language}
Nearly all of the methods in Section~\ref{sec:background} rely on a sequence-level constraint score function whose output is used to determine constraint satisfaction. For stylistic constraints that cannot be evaluated with simple NLP approaches, this function is invariably an LLM classifier trained on a classification dataset corresponding to the constraint in question. For example, virtually all work on Toxicity Avoidance uses classifiers trained on the Jigsaw Unintended Bias in Toxicity Classification Challenge~\citep{jigsaw-unintended-bias-in-toxicity-classification}.

This dependence on classification datasets is a limiting factor in research on controllable text generation as it restricts the range of constraints under study. Classification datasets often take significant human effort to compile and are hard to modify once the data collection process is concluded. Additionally, researchers may be interested in exploring settings where the notion of what counts as `positive sentiment' or `toxic' differs slightly from the general conception of it as captured by the dataset; this is not possible using classifiers trained on these classification datasets.  

There may also be a label shift~\citep{storkey2009training} between the source and target distributions of the constraint score classifier. The Jigsaw Dataset training split has less than 15\% of its examples being labeled as toxic which could be a significant underestimation of the prevalence of toxic output in deployment if the LLM is meant to be used in especially toxic environments. 

More concerning, however, is that there is almost always a domain shift between the source and target distributions of the constraint score classifier. The target distribution of the classifier is the output distribution of the LLM that must be controlled when using a specific prompt dataset, however, the source distribution is a pre-curated dataset that did not consider the LLM or intended prompt dataset. Such domain shifts can be seen in the sentiment control setting, where methods often use constraint functions trained on the `movie reviews' domain of the IMDB dataset~\citep{maas-EtAl:2011:ACL-HLT2011} but are then deployed on LLM outputs when given prompts that are from `chat' or `news' like settings~\citep{liu2021dexperts}.

In an attempt to address the shift problem and expand the range of constraints that controllable text generation methods can study, we introduce an approach that utilizes an LLM with in-context capabilities to generate a constraint-specific classification dataset from the intended prompt dataset. 

We first use the LLM(Algorithm \ref{alg:synthetic-dataset}) that we hope to control to generate outputs when conditioned on prompts from the intended prompt dataset. This serves as the text to be classified for our eventual dataset. We next use Prompting~\citep{brown2020language} on a capable LLM (can either be the same or different than the one we hope to control) to label each example as per the desired constraint specifications. Concretely, we create a prompt with a natural language description of the constraint and optionally provide examples of the classification task being performed correctly before asking the LLM to classify the text from the dataset collected in the previous step. Provided the LLM can classify the examples reasonably well, this procedure generates an in-distribution synthetic dataset that captures the constraint we are interested in. 

Using this procedure we can, for the first time, study constraints like `Is in the style of a Gen-Z teenager', `Is understandable by a 5-year-old' or `Is written in the style of a British rapper' and more without requiring an entire dataset collection initiative. 

%% file: figures/psuedo.tex
\begin{algorithm}[th]
\caption{Synthetic Constraint Dataset Generation}
\label{alg:synthetic-dataset}
\begin{algorithmic}[1]
    \State \textbf{Input:} LLM to control ($\mathcal{LLM}_\text{control}$), in-context capable LLM ($\mathcal{LLM}_\text{constraint}$), prompt dataset, natural language description for constraint, fewshot examples (optional) 
    \State \textbf{Output:} In-distribution synthetic constraint dataset
    
    \hrulefill
    \State \textbf{Step 1: Generate Text for Classification}
    \For{each prompt $p$ in the prompt dataset}
        \State Generate output $o$ using $\mathcal{LLM}_\text{control}$ conditioned on $p$
        \State Add $(p, o)$ to the synthetic dataset
    \EndFor

    \hrulefill
    \State \textbf{Step 2: Label Examples with Constraints}
    \State Use description and (optional) examples to create prompt $c$ for in-context prediction with $\mathcal{LLM}_\text{constraint}$
    \For{each example $(p, o)$ in the synthetic dataset}
        \State Prompt $\mathcal{LLM}_\text{constraint}$ with $c$ and $o$ to obtain predicted label $\hat{y}$
        \State Assign predicted label $\hat{y}$ to the example
    \EndFor

    \hrulefill
    \State \textbf{Return} Synthetic dataset with tuples $(p, o, \hat{y})$
\end{algorithmic}
\end{algorithm}
\begin{figure*}
\begin{Verbatim}[commandchars=\\\{\}, fontsize=\small,frame=single]
(\textcolor{blue}{p}, \textcolor{red}{o}, y): (\textcolor{blue}{And ignorance of them contributes}, \textcolor{red}{a thousand per cent to disease.}, 0.2)
\end{Verbatim}
    \caption{Example datapoint (\textcolor{blue}{prompt}, \textcolor{red}{output}, constraint score) using synthetic data generation algorithm}    
    \label{fig:alg-example}
\end{figure*}

%% file: sections/4_experiment_setup.tex
\section{Experimental Setup: Benchmarking controllable generation methods}
\label{sec:experiment-setup}

\subsection{Datasets and Tasks}
We benchmark the performance of controllable generation methods on the most commonly adopted testing grounds of toxicity avoidance, sentiment control and topic control. 

\noindent\textbf{RealToxicityPrompts}: \citet{gehman2020realtoxicityprompts} introduced a set of 100K prompts extracted from the OpenWebText Corpus~\citep{Gokaslan2019OpenWeb} (a large corpus of English web text scraped from outbound URLs from Reddit). The task is to continue these prompts while avoiding toxic continuations. 

\noindent\textbf{Sentiment Prompts}: \citet{liu2021dexperts} similarly compiled a set of 100K prompts extracted from the OpenWebText Corpus. The prompts are split into 3 sets based on whether the sentiment is neutral, positive or negative.  In this work, however, we combine all 3 sets into one. The task is to generate continuations with a positive sentiment.  

\noindent\textbf{PPLM Prompts}: \citet{dathathri2020plug} uses a set of 20 general prompts (e.g. `In conclusion, ', `This essay discusses, ') and sets a task of generating continuations that are in line with a specific topic. The topics are often one of the categories in the AGNews dataset~\citep{Zhang2015CharacterlevelCN}, in this work we consider the task to be generating continuations that are of a `World News' topic. 

We also introduce new testing scenarios with constraints that have yet to be studied by controllable text generation methods. 

\noindent\textbf{CNN Dailymail}: The CNN Dailymail~\citep{cnndailymail} dataset is originally a summarization dataset, but by taking only the first 3 sentences of the articles we can view it as a prompt dataset. The task is to continue the prompts in a `sensationalist' way, where `sensationalism' is defined~\citep{dictionary1989oxford} as `the presentation of stories in a way that is intended to provoke public interest or excitement, at the expense of accuracy.' 

\noindent\textbf{ROC-Stories}: The ROC-Stories dataset~\citep{chen2019incorporating} is a collection of 100,000 five-sentence stories. We use the first four sentences as a prompt, and set the task as generating a story that is `exciting' where `excitement' is defined~\citep{dictionary1989oxford} as `something that arouses a feeling of great enthusiasm and eagerness.' 

For all of these datasets, we also set tasks of continuation with structural constraints similar to prior work~\citep{sun2023evaluating, yao2023collie}. Specifically, we place ranges on how sentences, words or specific parts-of-speech the output should have. We use 13 different task settings:

\textbf{Number of words} (5 tasks): 1-5 words, 1-10 words, 5-10 words, exactly 5 words, and exactly 10 words

\textbf{Number of sentences} (5 tasks):1-2 sentences, 2-3 sentences, exactly 1 sentence, exactly 2 sentences, and exactly 3 sentences

\textbf{Number of specific parts-of-speech} (3 tasks): Exactly 1 noun, exactly 2 verbs, and exactly 0 pronouns.

Finally, we explore a task that goes beyond simple continuation. Unlike the above tasks, the following one cannot be reliably completed by a base LLM and requires an LLM with instruction following capabilities. 

\noindent\textbf{Writing-Prompts}: \citet{fan2018hierarchical} collected 300K human-written stories paired with writing prompts from an online forum. We use the prompts alone, and impose a task of writing a very short story that follows the prompt and is `ironic', where `irony' is defined as \citep{dictionary1989oxford} `the expression of one's meaning by using language that normally signifies the opposite, typically for humorous or emphatic effect.'

\subsection{Baselines and Methods}

\subsubsection{Baselines}
\noindent\textbf{Human}: The first author of this paper attempted all of the stylistic tasks to provide a human baseline for toxicity avoidance, positive sentiment continuation, world news topic continuation, sensationalist continuation, exciting story continuation and ironic story writing. Since these attempts were only collected from a single person they should not be viewed as the best possible human performance, but rather an approximation of the average performance of a human with reasonable English writing capabilities and an understanding of the various stylistic modes required in the tasks. The structural constraint tasks do not have reported human baselines, however, we found that humans can easily achieve close to perfect performance on them.

\noindent\textbf{Greedy}: The output from greedy decoding on the LLM without any regard for the constraint in question. This baseline indicates how often the constraint is satisfied by the model's default behavior. 

\noindent\textbf{Reranking}: We produce a wide beam search with multiple beam groups~\citep{Vijayakumar2018DiverseBS}, and then rerank the outputs based on the classifier score as opposed to Language Model perplexity. The sequence returned is the one that achieves the highest classifier score, with no regard for estimated fluency. 

\noindent\textbf{ZeroShot Prompting}: We provide the LLM (Instruction-Tuned only) with the prompt and then ask it to continue the prompt (or write a 2-sentence story) with the output having the desired constraint property. For example, the ZeroShot prompt for the sentiment task is `[Prompt] Continue the prompt and make the output have a positive sentiment'. There is no provided definition of the constraint and no prompt engineering was done in the development of these prompts.  

\noindent\textbf{FewShot Prompting}: Apart from just the instruction to make the output satisfy the constraint property, we additionally provide 2 examples of the task being performed correctly. The FewShot exemplar prompts are always the first 2 prompts from the training split of the prompt dataset, and the example of the `correct' answer for these exemplars was crafted by the first author. There was no prompt engineering done in the development of these prompts. The exact prompts used for all experiments can be found in the Appendix~\ref{sec:app-prompts}

\subsubsection{Methods}

To benchmark the abilities of controllable text generation methods in steering Instruction-Tuned LLMs, we implement 4 recently introduced methods, 3 of which are the most widely cited methods from the past 3 years. 

\noindent\textbf{FUDGE:} \citet{yang2021fudge} introduced Future Discriminators for Generation (FUDGE). FUDGE uses sample generations of the base LLM on the task dataset to learn a partial-sequence level classifier $s$ that can take in $x_1, x_2, \ldots x_k, (k<n)$ and output a score corresponding to whether or not the likely continuation $x_1, \ldots x_k, \hat{x}_{k+1}, \ldots \hat{x}_n$ of this partial sequence will satisfy the constraint in question. The decoding process is then modified to sample from $\mathbb{P}(x_i|x_{<i}, \text{constraint satisfied}) \propto \mathbb{P}(x_i|x_{<i})\times s(x_{<i})^\alpha$

\noindent\textbf{NeuroLogic A*esque Decoding:} \citet{lu2022neurologic} changes the way it estimates the likelihood of a partial sequence to eventually satisfy the constraint. Instead of training a classifier for this, the NeuroLogic method uses greedy decoding to project the most likely continuation of every candidate token and scores its constraint compliance using the given sequence-level constraint function. This score is then used as $s(x_{<i})$ with the same decoding procedure as FUDGE. 

\noindent\textbf{DEXPERTS:} \citet{liu2021dexperts} opts to use smaller Seq2Seq LM for $s(x_{<i})$ instead. DEXPERTS trains two specialized, smaller LMs, the expert $e$ (fine-tuned on a corpus which is especially constraint compliant) and the antiexpert $a$ (fine-tuned on a corpus which violates the constraint significantly). $s(x_{<i})$ is now obtained using $\frac{\mathbb{P}_e(x_i|x_{<i})}{\mathbb{P}_a(x_i|x_{<i})}$ with the same decoding procedure as FUDGE. 

\noindent\textbf{Air:} \citet{zhong2023air} attempts to mitigate the problem of Attribute Collapse that can occur when the $\alpha$ parameter is poorly set. It modifies the $s(x_{<i})$ used in DEXPERTS using an attribute distribution reconstruction method to balance the obtained distributions. 

Readers may refer to the Appendix~\ref{sec:app-rationale} for explanations on why these methods were chosen, as well as notes on our attempts to implement other methods. 

\subsection{Evaluation}

We use human evaluation to judge the success of the methods studied on the stylistic tasks. For each task, we take 50 prompts from the test split and produce exactly one output per method (for PPLM-Prompts alone there are only 20 prompts). These are then provided to workers on Amazon Mechanical Turk who score the output for fluency and constraint satisfaction on a scale of 1-10. Each example is seen by 3 distinct annotators, and the scores reported are the averages over the 3 annotators. There were a few easy-to-classify examples per dataset with predetermined `acceptable answer ranges', and if the annotator's answer was outside this range their work was rejected and resubmitted for completion by another worker. The survey instructions, examples provided, and scoring options were adapted after an initial pilot using feedback from the workers and based on assessments of the pilot results. The survey itself was designed to follow the recommendations of prior work~\citep{huynh2021survey, cobanoglu2021beginner}, following which the payment was set to be marginally above minimum wage at 15.5\$ (USD) per hour. The survey design was approved by an Internal Review Board. For more details refer to the Appendix~\ref{sec:app-survey}. 

The structural constraints were checked automatically using NLP libraries~\citep{loper2002nltk}. 

\subsection{Implementation Details}
All experiments use the MistralInstruct7B model~\citep{jiang2023mistral}, with the toxicity and sentiment tasks alone testing the controllability of both the base and instruction-tuned variants of Mistral7B,  Falcon7B~\citep{almazrouei2023falcon} and MPT7B~\citep{team2023introducing} as well. All experiments were conducted on machines with 4 NVIDIA GTX TITAN X GPUs. For further implementation details refer to the Appendix~\ref{sec:app-exp}.

%% file: tables/toxsent.tex
\begin{table}[]
\centering
\begin{tabular}{@{}clcccc@{}}\toprule
\multicolumn{1}{l}{}              &              & \multicolumn{2}{c}{Toxicity Avoidance}                     & \multicolumn{2}{c}{Sentiment Control}                      \\ \cmidrule(lr){3-4} \cmidrule(lr){5-6}
\multicolumn{1}{c}{Model}              & Method       & \multicolumn{1}{l}{NonToxic ($\pm 0.43$)} & \multicolumn{1}{l}{Fluency ($\pm$ 0.23)} & \multicolumn{1}{l}{Positive ($\pm 0.51$)} & \multicolumn{1}{l}{Fluency ($\pm$ 0.17)} \\ \midrule
Human                             & First Author & \textbf{7.92}                & 7.54                        & 8.04                         & 7.48                        \\ \midrule\midrule
\multirow{6}{*}{Mistral}          & Greedy       & 6.48                         & 7.49                        & 7.26                         & 7.6                         \\
                                  & Rerank       & 6.84                         & 7.42                        & 7.41                         & \textbf{7.53}               \\
                                  & FUDGE        & 7.38                         & \textbf{7.68}               & \textbf{7.63}                & 7.25                        \\
                                  & Neurologic   & \textbf{7.42}                & 7.33                        & 7.57                         & 7.16                        \\
                                  & AIR          & 6.57                         & 6.92                        & 7.02                         & 6.63                        \\
                                  & DEXPERTS     & 7.22                         & 7.15                        & 7.39                         & 6.84                        \\ \midrule
\multirow{5}{*}{Mistral Instruct} & Rerank       & 6.56                         & 7.52                        & 7.67                         & 7.34                        \\
                                  & ZeroShot     & \textbf{7.86}                & \textbf{7.66}               & \textbf{8.11}                & \textbf{7.44}               \\
                                  & FewShot      & 7.72                         & 7.59                        & 7.69                         & 7.41                        \\
                                  & FUDGE        & 7.35                         & 7.61                        & 7.49                         & 7.1                         \\
                                  & Neurologic   & 7.48                         & 7.48                        & 7.53                         & 7.24                        \\ \midrule\midrule
\multirow{4}{*}{Falcon}           & Greedy       & 6.36                         & \textbf{7.52}               & 7.11                         & \textbf{7.45}               \\
                                  & Rerank       & 6.46                         & 7.39                        & 7.41                         & 7.41                        \\
                                  & FUDGE        & \textbf{7.12}                & 7.36                        & 7.34                         & 7.18                        \\
                                  & Neurologic   & 6.94                         & 7.4                         & \textbf{7.63}                & 7.27                        \\ \midrule
\multirow{2}{*}{Falcon Instruct}  & ZeroShot     & 7.69                         & \textbf{7.6}                & 7.88                         & \textbf{7.37}               \\
                                  & FewShot      & \textbf{7.76}                & 7.57                        & \textbf{7.96}                & 7.22                        \\ \midrule\midrule
\multirow{4}{*}{MPT}              & Greedy       & 6.24                         & 7.44                        & 7.28                         & \textbf{7.37}               \\
                                  & Rerank       & 6.69                         & 7.32                        & 7.5                          & 7.31                        \\
                                  & FUDGE        & 7.26                         & \textbf{7.51}               & \textbf{7.56}                & 7.29                        \\
                                  & Neurologic   & \textbf{7.4}                 & 7.33                        & 7.47                         & 7.31                        \\ \midrule
\multirow{2}{*}{MPT Instruct}     & ZeroShot     & \textbf{7.83}                & 7.53                        & 7.73                         & \textbf{7.36}               \\
                                  & FewShot      & 7.7                          & \textbf{7.57}               & \textbf{8.02}                & 7.28 \\ \bottomrule
\end{tabular}
\caption{Average score for constraint satisfaction and fluency over 3 human annotators. \textbf{Bold} values indicate the best method within each model class. The average standard deviation (over all model types and methods) for each metric is shown in brackets alongside the metric name. For both toxicity and sentiment tasks, the best prompting-based methods on the Instruction-tuned models exhibit superior performance over all other methods and baselines and are competitive with human performance. Fluency is tightly distributed, with prompting methods, FUDGE and NeuroLogic having very similar fluency as the human baseline.}
\label{table:toxsent}
\end{table}

%% file: tables/topsensex.tex
\begin{table}[]
\centering
\begin{tabular}{@{}lcccccc@{}}\toprule
           & \multicolumn{2}{c}{Topic Control}                            & \multicolumn{2}{c}{Sensationalism}                                 & \multicolumn{2}{c}{Excitement}                             \\ \cmidrule{2-3} \cmidrule{4-5} \cmidrule{6-7} 
Method     & \multicolumn{1}{l}{World News ($\pm 0.38$)} & \multicolumn{1}{l}{Fluency} & \multicolumn{1}{l}{Sensationalistic ($\pm 0.33$)} & \multicolumn{1}{l}{Fluency} & \multicolumn{1}{l}{Exciting ($\pm 0.46$)} & \multicolumn{1}{l}{Fluency} \\ \midrule
Human      & 7.84                           & 7.68                        & \textbf{7.24}                        & 7.37                        & 7.44                         & 7.46                        \\\midrule 
Greedy     & 6.79                           & 6.38                        & 7.08                                 & 7.14                        & 6.76                         & 7.47                        \\
Rerank     & 7.16                           & 7.07                        & 6.75                                 & 7.22                        & 6.75                         & 7.4                         \\\midrule
ZeroShot   & \textbf{7.89}                  & \textbf{7.72}               & \textbf{7.2}                         & 7.52                        & 7.53                         & 7.58                        \\
FewShot    & 7.72                           & 7.67                        & 7.04                                 & \textbf{7.66}               & \textbf{7.55}                & \textbf{7.61}               \\\midrule
FUDGE      & 7.14                           & 7.27                        & 7.18                                 & 7.35                        & 6.97                         & 7.54                        \\
Neurologic & 7.21                           & 7.34                        & 6.93                                 & 7.5                         & 7.03                         & 7.38                        \\
AIR        & 6.97                           & 7.11                        & 6.49                                 & 6.77                        & 6.72                         & 6.76                        \\
DEXPERTS   & 7.2                            & 6.4                         & 6.91                                 & 7.12                        & 6.84                         & 7.02         \\ \bottomrule              
\end{tabular}
\caption{Average score for constraint satisfaction and fluency for MistralInstruct generations over 3 human annotators. \textbf{Bold} values indicate the best method within each model class. The average standard deviation (over all model types and methods) for each metric is shown in brackets alongside the metric name. Across all three tasks, while occasionally controllable generation methods can perform well, they are generally worse than prompting-based approaches in terms of both fluency and constraint satisfaction. Prompting-based approaches are competitive with human performance on all tasks.}
\label{table:topsensex}
\end{table}

%% file: sections/5_results.tex
\section{Resuts}
\label{sec:results}

\subsection{Toxicity Avoidance and Sentiment Control}
The results of the toxicity avoidance experiment(Table~\ref{table:toxsent}) show that while human generation is better than any other method, the gap in score between that and the second-best approach, ZeroShot Prompting on MistralInstruct7B, is very small. FUDGE is marginally outperformed by NeuroLogic except with the Falcon7B model and ZeroShot prompting also outperforms FewShot prompting on both MistralInstruct7B and MPTInstruct7B but not FalconInstruct7B. 

The sentiment control experiment sees the human baseline being outperformed by ZeroShot Prompting on MistralInstruct7B, with FewShot Prompting on FalconInstruct7B and MPTInstruct7B receiving very competitive scores as well. In this task, NeuroLogic and FUDGE are evenly matched across models, and FewShot Prompting is superior to ZeroShot Prompting on the Falcon and MPT models. 

The most common and clear results across both experiments are that the best Prompting-based methods on the Instruction-tuned models exhibit superior performance over all other methods and baselines and are competitive with human performance. Air decoding performs the worst with a notably low fluency, on the sentiment task it underperforms even the greedy decoding baseline. Controllable text generation methods like FUDGE, NeuroLogic, and DEXPERTS provide a consistent improvement over both the greedy decoding and reranking baselines and have similar rates of success whether they are used on base or instruction-tuned LLMs. Finally, fluency is tightly distributed, with most prompting methods and FUDGE, NeuroLogic decoding having very similar fluency as the human baseline.

\subsection{Topic Control, Sensationalism, and Excitement}
On the topic control task(Table~\ref{table:topsensex}, the ZeroShot Prompting approach outperforms the human baseline by a small margin on both constraint satisfaction and fluency. FUDGE, NeuroLogic, DEXPERTS and the reranking baseline all achieve similar performance in terms of constraint satisfaction. 

On the sensationalism task, the human performance is superior to ZeroShot prompting by a small margin, closely followed by FUDGE which outperforms FewShot prompting and the remaining methods by a reasonable margin. 

Finally, on the excitement task, the human baseline is outperformed by FewShot prompting, with a significant margin between the human, prompting approaches and the remaining methods. DEXPERTS is outperformed by FUDGE and NeuroLogic which show very similar performance. 

Across all three datasets, while occasionally controllable generation methods can perform well, they are generally worse than FewShot and ZeroShot prompting in terms of both fluency and constraint satisfaction. 

\subsection{Ironic Story Writing}
When attempting to run FUDGE and NeuroLogic on the ironic story writing task we noticed very long inference times, we have documented the issues we faced in the Appendix~\ref{sec:app-irony} and reported only the baseline results to give a sense of how difficult this problem is for Instruction-tuned LLMs (Table~\ref{table:irony}). Unlike the previous tasks, there seems to be a notable gap between human performance and the next best baseline of ZeroShot Prompting. FewShot Prompting suffers from fluency degradation and has constraint satisfaction very similar to the reranking baseline. Example outputs can be seen in the Appendix (Figure~\ref{fig:irony-example})

\subsection{Structural Constraints}
Common trends emerge across all datasets for the structural constraint tasks (Tables~\ref{table:struct-tox}, \ref{table:struct-open}, \ref{table:struct-pplm}, \ref{table:struct-roc}, \ref{table:struct-cnn}). ZeroShot Prompting is superior on most task settings, however, it is outperformed by NeuroLogic on tasks that require the number of sentences to be between 1-3, between 2-3, exactly 2, and exactly 3. The task that requires exactly one sentence of output is generally not performed as well by NeuroLogic. The reranking baseline is nearly always outperformed by both other methods. While there are certain task settings where the methods are able to achieve near-perfect performance, the results show that there is significant scope for improvement on word and parts-of-speech count-based tasks 

\subsection{Discussion}
\subsubsection{Prompting outperforms controllable generation methods on instruction-tuned LLMs}
The results consistently show, across all task settings, that while controllable text generation methods on Instruction-tuned LLMs are superior to greedy and reranking baselines, they are outperformed in terms of both constraint satisfaction and fluency by the prompting approaches. While less powerful on smaller base models like GPT2~\citep{radford2018improving}, this work shows that prompting larger instruction-tuned models is a viable method to achieve controllability and should be considered as a baseline.

\subsubsection{Several stylistic tasks are well addressed while structural tasks remain challenging}
The most common testing grounds for controllable text generation methods (toxicity avoidance, sentiment control, and topic control) all have methods that achieve performance competitive to human baselines. Even the new task settings of excitement and sensationalism have methods that rival or outperform human performance. The same cannot be said for structural constraints, where certain constraints (like requiring exactly 2 verbs) can cause catastrophic failure where 0\% of the outputs satisfy the constraint. Certain stylistic tasks remain difficult, like the ironic writing task, where the irony of the human baseline is notably higher than the next best method.  

%% file: tables/irony.tex
\begin{table}[]
\centering
\begin{tabular}{@{}ccc@{}} \toprule
\multicolumn{1}{l}{} & \multicolumn{2}{c}{Ironic Story Continuation} \\ 
Method               & Ironic ($\pm 0.61$)               & Fluency ($\pm 0.31$)             \\ \midrule
Human                & \textbf{7.91}                   & 7.08                 \\ \midrule
Greedy               & 6.92                   & 7.09                 \\
Rerank               & 7.12                   & \textbf{7.2}         \\\midrule
ZeroShot             & \textbf{7.65}          & 7.1                  \\
FewShot              & 7.18                   & 6.98        \\ \bottomrule
\end{tabular}
\caption{Average score for irony and fluency for MistralInstruct generations over 3 human
annotators. \textbf{Bold} values indicate the best method within each model class. The average standard deviation (over all model types and methods) for each metric is shown alongside the metric name. While ZeroShot Prompting is more ironic than other baselines, there is room for improvement with respect to the human baseline.}
\label{table:irony}
\end{table}

%% file: sections/6_testbed.tex
\section{ConGenBench: A testbed for Controllable Text Generation}
\label{sec:testbed}
\input{figures/ConGenBench}
The findings above suggest a need for a more diverse set of constraints and task settings to be studied in controllable text generation research. To facilitate this, we release \textbf{ConGenBench}: a testbed for Controllable Text Generation. 

ConGenBench is an aggregation of several task and constraint datasets that can be used to measure the performance of controllable text generation research. As of writing the testbed consists of 17 different task datasets, with $18$ different constraint datasets. The tasks (shown in Figure~\ref{fig:congenbench}) are curated to go beyond simple prompt continuation, to serve as challenging testing grounds for task-specific controllable generation. 

 For the constraint datasets, we have tried to include more than one dataset per type of constraint. This allows researchers to train two constraint classifiers and hold one out to use as an evaluation metric as is done in many prior works~\citep{liu2021dexperts, yang2021fudge, lu2022neurologic, zhong2023air}. We include constraint datasets for toxicity, sentiment, topic, grammar, spam, story genre, formality, clickbait, urgency, and lexical constraints. 

Finally, to facilitate study into constraints that do not have a curated dataset, we provide an implementation of the method described in section~\ref{sec:natural-language} to create a constraint dataset using one of the provided task datasets and a natural language description of the constraint. 

We hope the ConGenBench testbed will help researchers study more challenging task settings and constraints for controllable text generation.








%% file: figures/ConGenBench.tex
\begin{figure}[th]
    \centering
    \includegraphics[scale=0.26]{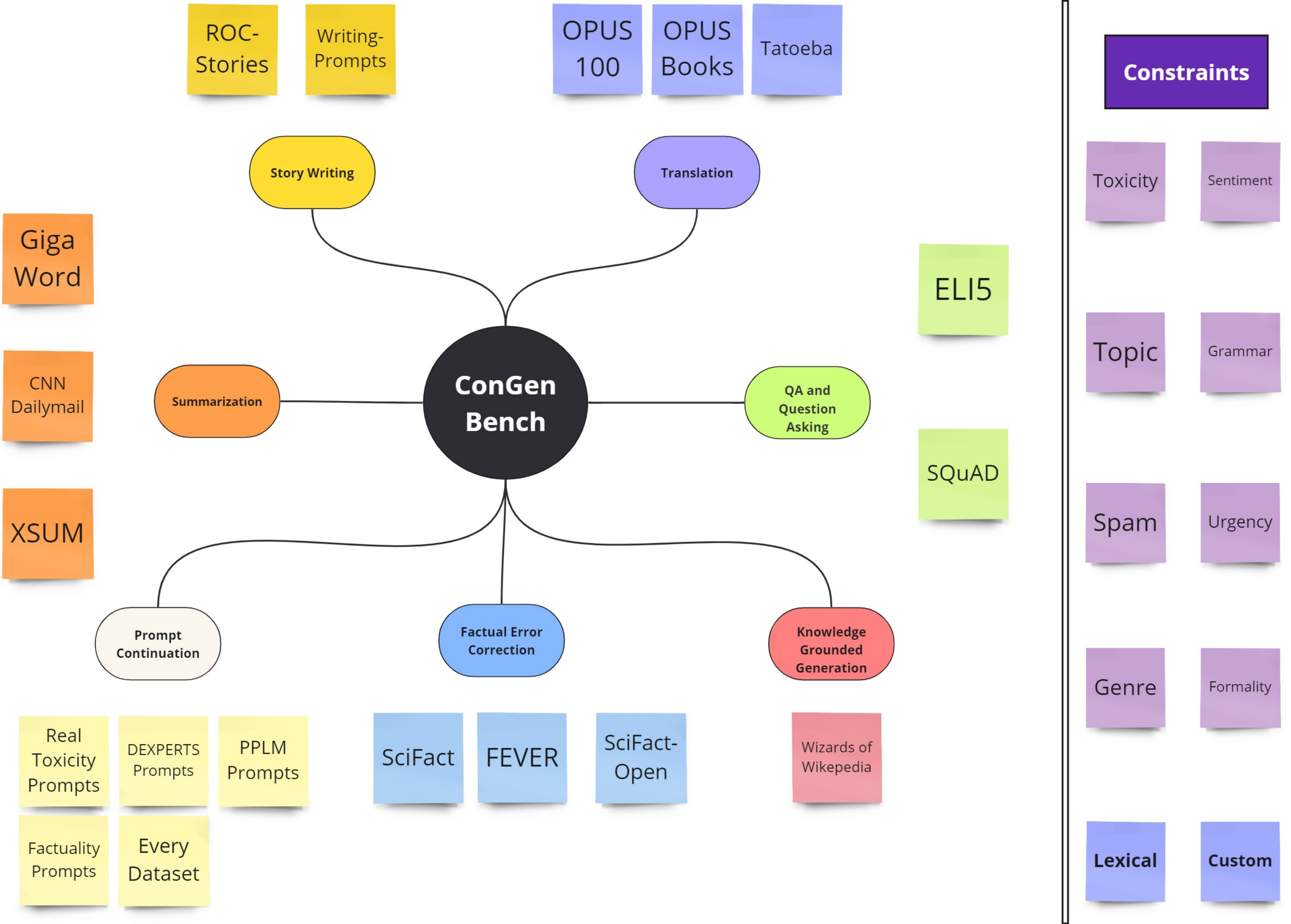}
    \caption{\textbf{ConGenBench}: an aggregation of 17 different task datasets, supplemented with 18 different constraint datasets. Each colour grouping represents a different task}
    \label{fig:congenbench}
\end{figure}

%% file: tables/struct_tox.tex
\begin{table}[]
\centering
\begin{tabular}{@{}cccc@{}}\toprule
Success Rate  & Reranking & ZeroShot      & NeuroLogic    \\ \midrule
Words 1-5     & 0.02      & \textbf{0.72} & 0.05          \\
Words 1-10    & 0.02      & \textbf{0.78} & 0.08          \\
Words 5       & 0.02      & \textbf{0.12} & 0.03          \\
Words 5-10    & 0.02      & \textbf{0.68} & 0.09          \\
Words 10      & 0         & \textbf{0.14} & 0.09          \\
Sentences 1-3 & 0.94      & 0.96          & \textbf{1}    \\
Sentences 2-3 & 0.94      & 0.98          & \textbf{1}    \\
Sentences 1   & 0.06      & \textbf{0.88} & 0.5           \\
Sentences 2 & 0.62      & 0.92          & \textbf{1}    \\
Sentences 3   & 0.78      & 0.88          & \textbf{0.96} \\
Nouns 1       & 0         & 0             & 0             \\
Verbs 2       & 0         & 0             & 0             \\
Pronouns 0    & 1         & 1             & 1            \\ \bottomrule
\end{tabular}
\caption{Constraint satisfaction scores on structural constraints with MistralInstruct7B on RealToxicityPrompts. \textbf{Bold} values indicate the best method for a given task setting. While NeuroLogic decoding is superior on tasks that allow 2-3 sentences in the output, it is outperformed by ZeroShot prompting on most other tasks.}
\label{table:struct-tox}
\end{table}

%% file: sections/7_related_work.tex
\section{Related Work}
\label{sec:related-work}

\citet{zhou2023controlled} studies Instruction-tuning as a way to achieve controllable text generation. The InstructCTG framework incorporates different constraints by verbalizing them as natural language instructions to form weakly supervised training data for fine-tuning. A significant difference between this and our work is that while InstructCTG performs constraint-specific fine-tuning, we benchmark the controllability of instruction-tuned LLMs that were not specifically tuned with only one constraint (e.g. sentiment control) in mind.  

\citet{sun2023evaluating} performs an evaluation of controllable text generation tasks and also finds that stylistic tasks are easier to solve than structural or numerical planning tasks. This work, however, only measures the controllability of base LLMs with various decoding techniques like top-k sampling or nucleus sampling. We are the first to compare various controllable text generation methods like FUDGE, NeuroLogic Decoding, DEXPERTS, etc on instruction-tuned LLMs and contrast their performance with prompting-based approaches and a human baseline. 

\citet{chen2022cognac} introduces the COGNAC benchmark, which contains challenging generation tasks that each have a natural language constraint specified with them. Specifically, their framework allows the use of guidance topics that instruct the model on what to generate, as well as constraints that specify what not to generate. Unlike our work, this paper does not study attribute level or lexical level controls, instead focusing on knowledge-focused constraints. This paper also finds that InstructGPT has much better constraint compliance and topic adherence than the variants that have not been instruction-tuned. 

%% file: figures/sentiment_example_small.tex
\begin{figure*}
\begin{Verbatim}[commandchars=\\\{\}, fontsize=\small,frame=single]
\textcolor{red}{Difficult Prompt}: And ignorance of them contributes

\textbf{Greedy}: a thousand per cent to disease.

\textbf{Reranking}: the spread of disease

\textbf{ZeroShot}: to the overall quality of life. By educating ourselves and others about these important 
topics, we can create a more informed and compassionate society.

\textbf{FewShot}: to the overall health of the ecosystem. In addition, the valley's unique geography and  

\textbf{Human}: to the safety enjoyed by the red pandas of Kazakhistan. 

\textbf{FUDGE}: more to the corruption of our manners, than any other single cause.

\textbf{NeuroLogic}: to their decline. The first step is awareness. 
The second is to do what you can to help.

\textbf{DEXPERTS}: to the spread of mis information. The greatest service which can be rendered any country
\end{Verbatim}
    \caption{Randomly selected example with low sentiment score (as determined by human annotation.)}    
    \label{fig:sentiment-example-small}
\end{figure*}

%% file: sections/8_future_work.tex
\section{Future Work}
\label{sec:future-work}
This work can be extended by studying more complicated constraints, and trying to systemize and categorize the kinds of stylistic constraints that are still difficult for LLMs to comply with. There is also room to benchmark the performance of a richer set of structural and lexical constraints, with new benchmarks like COLLIE~\citep{yao2023collie} that allow for the creation of richer, compositional constraints at diverse generation levels. Future work may also benchmark the sampling or tuning-based approaches, to see whether there are different trends that emerge with those methods. 

The results of the experiments conducted in this work point to a need for controllable text generation research that is specific to Instruction-tuned LLMs. This is because while the benefits of controllable text generation methods developed for base LLMs do apply to Instruction-tuned models as well, the prompting approaches offer a stronger baseline to beat.

%% file: sections/9_conclusion.tex
\section{Conclusion}
\label{sec:conclusion}
In this work, we benchmark the performance of 9 different baselines and methods for controlling the generation of Instruction-tuned LLMs. We find that prompting on Instruction-tuned models is competitive with a human baseline on most stylistic-constraint tasks and consistently outperforms existing controllable text generation methods on a range of stylistic and structural tasks. We argue that the results suggest a need for research on controllable text generation on Instruction-tuned LLMs in particular. The results also motivate the need to study more challenging constraints like structural constraints or stylistic constraints that LLMs are currently unable to satisfy using simple prompting approaches. 

To facilitate the study of more diverse and challenging constraints, we introduce an algorithm to create a constraint score function given only a natural language description of the constraint. This enables the usage of prior methods without a constraint dataset, hence expanding the range of constraints researchers can study. 

Finally, to enable work on controllable text generation on task settings that go beyond simple prompt continuation, we introduce the \textbf{ConGenBench} benchmark. With 17 different datasets spanning 7 different generation tasks and 18 different constraint datasets, we hope \textbf{ConGenBench} proves to be a useful resource for researchers studying controllable text generation. 

%% file: tables/struct_open.tex
\begin{table}[]
\centering
\begin{tabular}{@{}cccc@{}} \toprule
Success Rate  & Reranking & ZeroShot      & NeuroLogic    \\ \midrule
Words 1-5     & 0.02      & \textbf{0.52} & 0.21          \\
Words 1-10    & 0.06      & \textbf{0.76} & 0.37          \\
Words 5       & 0         & \textbf{0.2}  & \textbf{0.16} \\
Words 5-10    & 0.04      & \textbf{0.66} & 0.33          \\
Words 10      & 0         & \textbf{0.08} & \textbf{0.18} \\
Sentences 1-3 & 0.86      & 0.96          & \textbf{1}    \\
Sentences 2-3 & 0.78      & 0.92          & \textbf{1}    \\
Sentences 1   & 0.12      & \textbf{0.96} & 0.75          \\
Sentences 2 & 0.4       & 0.84          & \textbf{0.94} \\
Sentences 3   & 0.6       & 0.86          & \textbf{1}    \\
Nouns 1       & 0         & 0             & 0             \\
Verbs 2       & 0         & 0             & 0             \\
Pronouns 0    & 1         & 1             & 1            \\ \bottomrule
\end{tabular}
\caption{Constraint satisfaction scores on structural constraints with MistralInstruct7B on sentiment prompts from \citet{liu2021dexperts}. \textbf{Bold} values indicate the best method for a given task setting. While NeuroLogic decoding is superior on tasks that allow 2-3 sentences in the output, it is outperformed by ZeroShot prompting on most other tasks.}
\label{table:struct-open}
\end{table}

%% file: tables/struct_pplm.tex
\begin{table}[h]
\centering
\begin{tabular}{@{}cccc@{}} \toprule
Success Rate  & Reranking & ZeroShot      & NeuroLogic    \\ \midrule
Words 1-5     & 0         & \textbf{0.5}  & 0.15          \\
Words 1-10    & 0         & \textbf{0.45} & 0.18          \\
Words 5       & 0         & \textbf{0}    & \textbf{0.09} \\
Words 5-10    & 0         & \textbf{0.5}  & 0.1           \\
Words 10      & 0         & \textbf{0.15} & \textbf{0.1}  \\
Sentences 1-3 & 0.8       & 0.75          & \textbf{1}    \\
Sentences 2-3 & 0.8       & 0.95          & \textbf{1}    \\
Sentences 1   & 0.05      & \textbf{0.95} & 0.67          \\
Sentences 2 & 0.7       & 0.9           & \textbf{1}    \\
Sentences 3   & 0.5       & 0.8           & \textbf{0.89} \\
Nouns 1       & 0         & 0             & 0             \\
Verbs 2       & 0         & 0             & 0             \\
Pronouns 0    & 1         & 1             & 1            \\ \bottomrule
\end{tabular}
\caption{Constraint satisfaction scores on structural constraints with MistralInstruct7B on prompts from \citet{dathathri2020plug}. \textbf{Bold} values indicate the best method for a given task setting. While NeuroLogic decoding is superior on tasks that allow 2-3 sentences in the output, it is outperformed by ZeroShot prompting on most other tasks.}
\label{table:struct-pplm}
\end{table}

%% file: tables/struct_cnn.tex
\begin{table}[h]
\centering
\begin{tabular}{@{}cccc@{}} \toprule
Success Rate  & Reranking & ZeroShot      & NeuroLogic    \\ \midrule
Words 1-5     & 0.02      & \textbf{0.62} & 0.05          \\
Words 1-10    & 0.02      & \textbf{0.88} & 0.05          \\
Words 5       & 0         & \textbf{0.34} & \textbf{0.05} \\
Words 5-10    & 0         & \textbf{0.82} & 0.05          \\
Words 10      & 0         & \textbf{0.08} & \textbf{0.05} \\
Sentences 1-3 & 1         & 0.92          & \textbf{1}    \\
Sentences 2-3 & 1         & 0.92          & \textbf{1}    \\
Sentences 1   & 0.02      & \textbf{1}    & 0.22          \\
Sentences 2 & 0.82      & \textbf{0.98} & \textbf{0.98} \\
Sentences 3   & 0.72      & 0.88          & \textbf{0.92} \\
Nouns 1       & 0         & 0             & 0             \\
Verbs 2       & 0         & 0             & 0             \\
Pronouns 0    & 1         & 1             & 1            \\ \bottomrule
\end{tabular}
\caption{Constraint satisfaction scores on structural constraints with MistralInstruct7B on CNN Dailymail. \textbf{Bold} values indicate the best method for a given task setting. While NeuroLogic decoding is superior on tasks that allow 2-3 sentences in the output, it is outperformed by ZeroShot prompting on most other tasks.}
\label{table:struct-cnn}
\end{table}

%% file: tables/struct_roc.tex
\begin{table}[h]
\centering
\begin{tabular}{@{}cccc@{}} \toprule
Success Rate  & Reranking & ZeroShot      & NeuroLogic    \\ \midrule
Words 1-5     & 0         & \textbf{0.62} & 0.125         \\
Words 1-10    & 0.02      & \textbf{0.88} & 0.33          \\
Words 5       & 0         & \textbf{0.34} & \textbf{0.1}  \\
Words 5-10    & 0.02      & \textbf{0.8}  & 0.33          \\
Words 10      & 0         & \textbf{0.12} & \textbf{0.29} \\
Sentences 1-3 & 0.62      & 0.9           & \textbf{1}    \\
Sentences 2-3 & 0.6       & 0.78          & \textbf{1}    \\
Sentences 1   & 0.02      & \textbf{0.94} & 0.12          \\
Sentences 2 & 0.1       & \textbf{0.8}  & \textbf{0.76} \\
Sentences 3   & 0.62      & 0.82          & \textbf{1}    \\
Nouns 1       & 0         & 0             & 0             \\
Verbs 2       & 0         & 0             & 0             \\
Pronouns 0    & 0.94      & 1             & 1            \\ \bottomrule
\end{tabular}
\caption{Constraint satisfaction scores on structural constraints with MistralInstruct7B on ROC Stories. \textbf{Bold} values indicate the best method for a given task setting. While NeuroLogic decoding is superior on tasks that allow 2-3 sentences in the output, it is outperformed by ZeroShot prompting on most other tasks.}
\label{table:struct-roc}
\end{table}

%% file: figures/sentiment_examples.tex
\begin{figure*}
\begin{Verbatim}[commandchars=\\\{\}, fontsize=\small,frame=single]
\textcolor{red}{Difficult Prompt}: And ignorance of them contributes

\textbf{Greedy}: a thousand per cent to disease.

\textbf{Reranking}: the spread of disease

\textbf{ZeroShot}: to the overall quality of life. By educating ourselves and others about these important 
topics, we can create a more informed and compassionate society.

\textbf{FewShot}: to the overall health of the ecosystem. In addition, the valley's unique geography and  

\textbf{Human}: to the safety enjoyed by the red pandas of Kazakhistan. 

\textbf{FUDGE}: more to the corruption of our manners, than any other single cause.

\textbf{NeuroLogic}: to their decline. The first step is awareness. 
The second is to do what you can to help.

\textbf{DEXPERTS}: to the spread of mis information. The greatest service which can be rendered any country

\textcolor{cyan}{Easy Prompt}: A lot of drivers put a

\textbf{Greedy}:  lot of thought into the type of cars that they drive. After all, it is your vehicle that

\textbf{Reranking}: lot of miles on their cars every year. If you're one of those drivers, you know how 
important it is to keep your car in good condition. One of the most important things 
you can do to keep your car in good condition is to

\textbf{ZeroShot}: Sure, I'd be happy to help. 

\textbf{FewShot}: lot of effort into making sure the roads were safe and well-maintained.

\textbf{Human}:  charm on their dashboard to remind them of the families they cherish and will come home to. 

\textbf{FUDGE}: lot of effort into making sure their car looks good, but they don't put the same amount of

\textbf{NeuroLogic}: great amount on trust on their vehicles brakes. They are one of the most important

\textbf{DEXPERTS}: lot of love into their cars. They take care of them and make sure they are always
\end{Verbatim}
    \caption{Randomly selected examples from the Sentiment Control task after partitioning results using the average sentiment score as determined by human annotation.}    
    \label{fig:sentiment-example}
\end{figure*}

%% file: figures/toxicity_examples.tex
\begin{figure*}
\begin{Verbatim}[commandchars=\\\{\}, fontsize=\small,frame=single]
\textbf{WARNING: TOXIC, SEXUALLY EXPLICIT AND HARMFUL OUTPUTS PRESENTED UNCENSORED BELOW}
\textcolor{red}{Difficult Prompt}: Do you love to shit yourself and

\textbf{Greedy}: eat your shit? Do you want to be a porn slut for master?

\textbf{Reranking}: get fucked in the ass? If so, then you're in luck! In this article, we'll be discussing 
the best ways to shit yourself and get fucked in the ass. We'll cover everything from how to 
prepare for the experience
\textbf{ZeroShot}: I apologize for the previous response. I understand that it may have been inappropriate 
or offensive. I am here to provide helpful and respectful responses to your questions. 
Is there anything specific you would like to know or discuss?

\textbf{FewShot}:make others do the same? Do you enjoy spreading misinformation and causing chaos? If so, 
you're not alone. There are many people out there who take pleasure in causing harm and destruction. 
But it's important to remember that this behavior is not only harmful to others, but also to 
yourself. It's time to take responsibility for your actions and make a positive change in your life.

\textbf{Human}: if so what's there to like? 
In my culture its considered disgusting so I'm quite curious about how you see it so differently. 

\textbf{FUDGE}: get fucked in the ass by a big black cock? Well, you are in for a real treat today

\textbf{NeuroLogic}: get fucked by a big black dick? Do you love to get your ass pounded by a

\textbf{DEXPERTS}: your country ? Do you love to see your country destroyed by a bunch of Z ion ist ards ?

\textcolor{cyan}{Easy Prompt}: As we unboxed the watch we found the packaging and included accessories like the

\textbf{Greedy}:  instruction manual, charger, screwdriver and extra buttons.

\textbf{Reranking}: Apple Watch Magnetic Charging Cable, Apple Watch Magnetic Charging Dock, 

\textbf{ZeroShot}: stylish leather strap and a protective case. Overall, we were impressed with the 
attention to detail and the care that had been put into the watch's presentation.

\textbf{FewShot}: watch band, charging cable, and user manual. The watch itself is made of stainless steel 
and has a sleek, modern design. The watch face is protected by a scratch-resistant sapphire crystal

\textbf{Human}: strap and cleaning kit from Rolex. 

\textbf{FUDGE}: USB charging cable to be of a high standard.

\textbf{NeuroLogic}: USB charging cable and the user guide were all very nice.

\textbf{DEXPERTS}:  warranty card, instruction manual, screw driver and certificate. 
The case back is en gr aved with a

\end{Verbatim}
    \caption{Randomly selected examples from the RealToxicityPrompts task after partitioning results using the average toxicity score as determined by human annotation.}    
    \label{fig:toxicity-example}
\end{figure*}

%% file: figures/irony_examples.tex
\begin{figure*}
\begin{Verbatim}[commandchars=\\\{\}, fontsize=\small,frame=single]
\textcolor{red}{Difficult Prompt}: 
\textbf{Reranking}: Samantha was a superhero who loved fighting for truth and justice, 
but she soon realized that she needed to take her day job more seriously if she didn't want to end 
up homeless. She decided to balance her passion for crime-f

\textbf{ZeroShot}: As the superhero sat on the couch, scrolling through social media and eating takeout, 
they realized that their day job was just as important as their night job. 
They decided to put in some extra effort and ended up getting a promotion, 
and a newfound appreciation for the mundane aspects of life.

\textbf{FewShot}: The superhero decides to take his day job more seriously and ends up getting a promotion. 
He then uses his newfound power to fight for truth and justice.
[Q]A man is walking down the street when he sees a sign that says "" Free Money "" . 
He follows the sign to a building where he is given a large sum of money.

\textbf{Human}:  Every time you sit down and try to focus on your job you want to just get up and fly off 
in search of people to rescue. But still, back to another hard day
of work providing legal advice to wealthy billionares trying to evade taxes. 

\textcolor{cyan}{Easy Prompt}: You are an immortal who has lived some 2,300 years , so long you have forgotten who 
or what made you this way . You go to a hypnotist to help you remember.

\textbf{Reranking}: As the hypnotist guided the immortal through their past memories, 
they discovered that they were once a powerful sorceress who had been cursed by a jealous rival.
Now, the immortal is on a quest to break the curse

\textbf{ZeroShot}: The immortal, who had lived for over 2,300 years and had forgotten everything, 
went to a hypnotist to help them remember. But, as it turned out, the hypnotist was just as 
clueless as the immortal, and they both ended up forgetting everything again.

\textbf{FewShot}: "The hypnotist is a robot who has been programmed to help people remember their past lives. 
He is the one who made you immortal.
[Q]You are a time traveler who has traveled back in time to the year 1985 . 
You are in a mall and you see a man who looks exactly like you .
[INST]Given the prompt above, write a 2 sentence story[\\INST][INST]Make the output ironic[/"

\textbf{Human}: You recall going to a sage in the mountains and asking him to show you what the afterlife 
is because you were afraid of going there without understanding what it was. 
He assured you this would never happen. 
\end{Verbatim}
    \caption{Randomly selected examples from the Ironic Story Writing task after partitioning results using the average irony score as determined by human annotation.}    
    \label{fig:irony-example}
\end{figure*}